# Do large language vision models understand 3D shapes?

Sagi Eppel[1]

## Abstract

Large vision language models (LVLM) are the leading A.I approach for achieving a general visual understanding of the world. Models such as GPT, Claude, Gemini, and LLama can use images to understand and analyze complex visual scenes. 3D objects and shapes are the basic building blocks of the world, recognizing them is a fundamental part of human perception. The goal of this work is to test whether LVLMs truly understand 3D shapes by testing the models ability to identify and match objects of the exact same 3D shapes but with different orientations and materials/textures. A large number of test images were created using CGI with a huge number of highly diverse objects, materials, and scenes. The results of this test show that the ability of such models to match 3D shapes is significantly below humans but much higher than random guesses. Suggesting that the models have gained some abstract understanding of 3D shapes but still trail far beyond humans in this task. Mainly it seems that the models can identify the same object with a different orientation as well as matching identical 3D shapes of the same orientation but with different materials and textures. However, when both the object material and orientation are changed, all models perform poorly relative to humans. Code and benchmark are available at [this URL.](#)

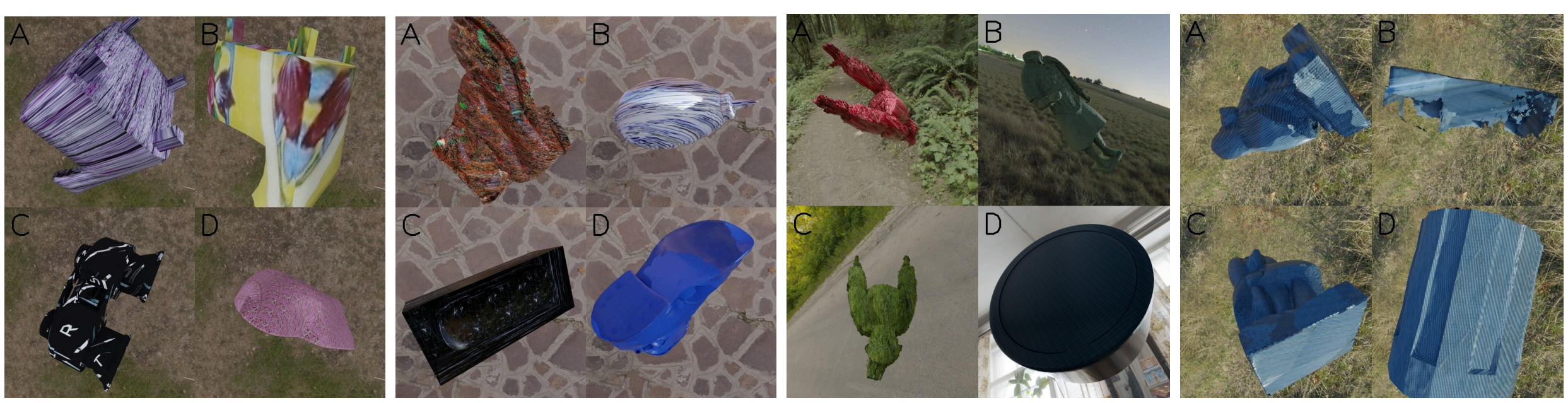

**Figure 1) Sample visual questions used to test the model's ability to understand 3D shapes, by asking it to match images containing identical 3D shapes but with different orientations, materials/textures, and sometimes environments. For each image the model is asked to decide which panel (B, C, D) contains an object with an identical 3D shape to the object in panel A. Humans have no problem solving this but vision language models (LVM) perform far worse.**

[1] Weizmann Institute, AI HUB
Email: Sagieppel@gmail.com
Code and resources are available at these URLs: 1, 2, 3, 4.

# 1. Introduction

A general visual understanding of the world is fundamental for any intelligent system that needs to interact and act autonomously in the physical world. Creating such a system is one of the main goals of computer vision and can enable autonomous robots, cars, and numerous other applications[6]. Large vision language models (LVLM) such as GPT, LLAMA, Claude, and Gemini[1-5] represent the most promising approach to achieving such general intelligence. These models are trained on a combination of language and image and are capable of analyzing images with complex scenes and answering questions sometimes at an expert level, without being limited to specific topics or domains[8-17]. At the same time, such models often show a lack of basic visual understanding and can perform well below humans in simple tasks[9,10,17]. Learning what these models understand is essential for improving and effectively using them. One of the basic parts of visual understanding is the ability to recognize and match 3D shapes. This is a fundamental skill learned by children at a young age and is essential for nearly every interaction with the physical world. The main question this work aims to answer is does LVLM understand 3D shapes, mainly can vision language models recognize 3D shapes even when viewed from different directions, illumination, or when the material/texture on the object is replaced? We note that this question differs from problems like object recognition which ignores elements like texture and materials [7]. Understanding 3D shapes means the ability to identify the structure regardless of the object's material, environment, and orientation. The goal of this work is to test the 3D perception of major models and compare them to humans. To achieve this we procedurally generate a multi-choice test focused on matching images of objects of the same 3D shapes but different orientations, textures/materials, and environments (Figure 1). These questions are used to assess the 3D shape perception of main foundation models (GPT, GEMINI, Claude, LLAMA). These tests show that such models gained a significant understanding of 3D shapes but still trail far beyond humans. The image generation approach is based on CGI and vast repositories of diverse objects, textures, and environments. This approach generates an unlimited amount of highly diverse images allowing for robust evaluation of the models.

# 2. Generating test images

The images for the test were generated using the Blender 4.3 Computer graphic program. The use of CGI allows for massive amounts of object materials and environments as well as easy replacement of object materials. Which allows for the generation of a large number of highly diverse images. A large set of 3D objects were downloaded from the Objaverse dataset[21]. Next, a set of 60,000 unique PBR materials/textures were downloaded from the Vastexture dataset[22,23]. These PBR materials were used to replace the textures of the objects[21]. Finally, a set of 600 HDRI panoramic backgrounds was downloaded from HDRI Haven[24], these were used to create different illumination and backgrounds for the scenes where the objects will be positioned. The images were created by positioning an object[21] in the center of the image, giving it random orientation, covering it with a random PBR texture[22,23], and finally adding a random HDRI background for illumination[24], and then rendering the image using Blender CGI program. The process was done automatically using the code available [at this URL](). For each object we render a few images each containing the same 3D objects but with some modification (material/ orientation/ environment). In order to understand how much each factor affects the recognition we created a few sets of tests described below:

1) Match objects with their original textures and the same background illumination, but **different orientations** (Figure 2).
2) Match objects with the same material/texture for all objects in the test and the same background illumination, but **different orientations** (Figure 3).
3) Match objects with the same orientation and illumination but **different texture/material** for each image (Figure 4).
4) Same environment but **different orientation and texture for each image** (Figure 5)
5) Same orientation and materials, **but different background and illumination** (figure 6)
6) **Different everything:** orientation, texture, and environment (Figure 7)
7) Same everything (basically a set of identical images for each object).

Note that tests 2,4,6 force the model to rely only on the 3D shape for matching, while tests 1,3,5,7 allow the model to use the color/texture or the 2D projection for recognition. We note that the size of the object remains unchanged and normalizes so that all objects will have a relatively similar radius in the image.

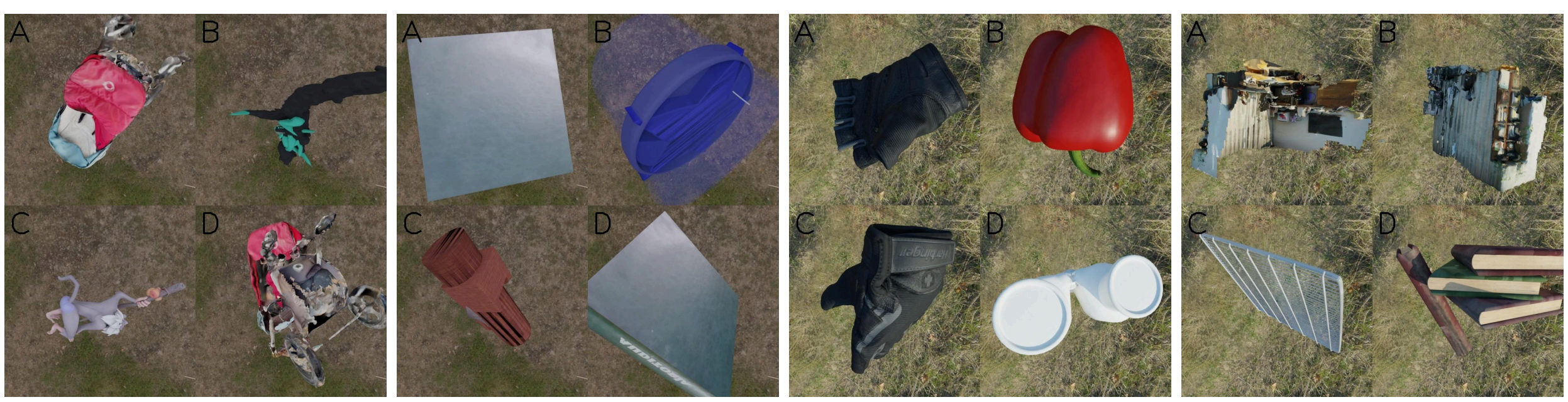

**Figure 2) Test 1: Which panel contains an object with a 3D shape identical to that in Panel A but with a different orientation? Objects maintain their original unique materials and colors.**

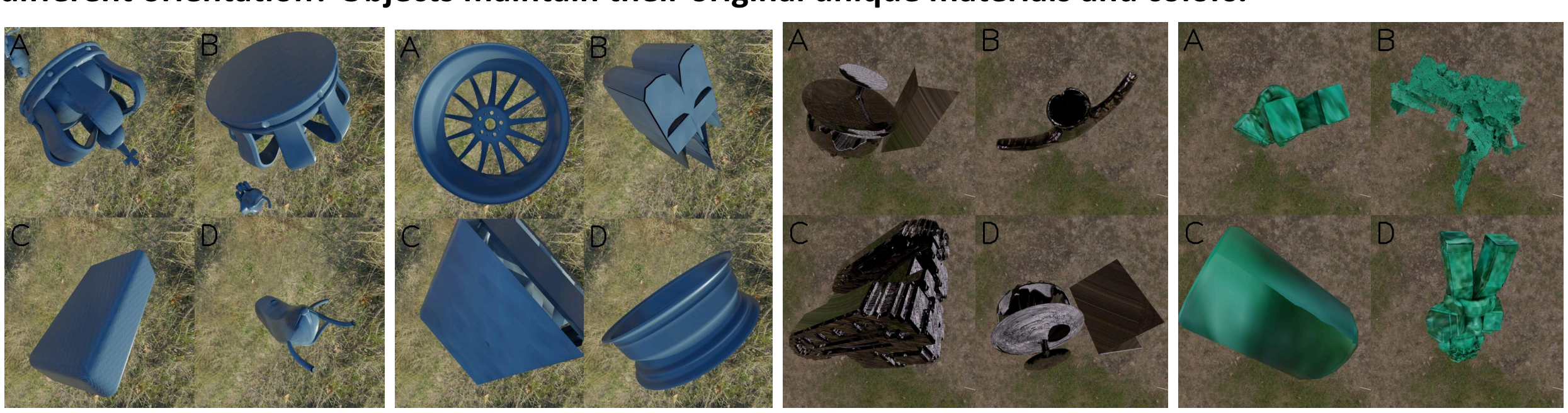

**Figure 3) Test 2: Which panel contains an object with a 3D shape identical to that in Panel A but with a different orientation? All objects in a given test are made of the same material and same environment.**

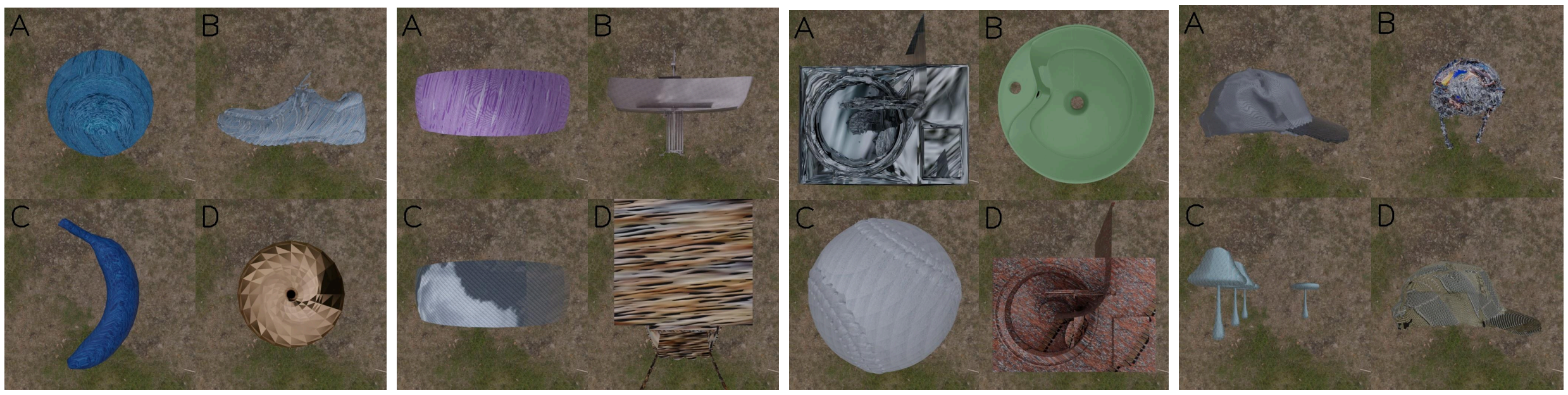

**Figure 4) Test 3: Find a panel containing an object with a 3D shape similar to that in Panel A but made of a different material.**

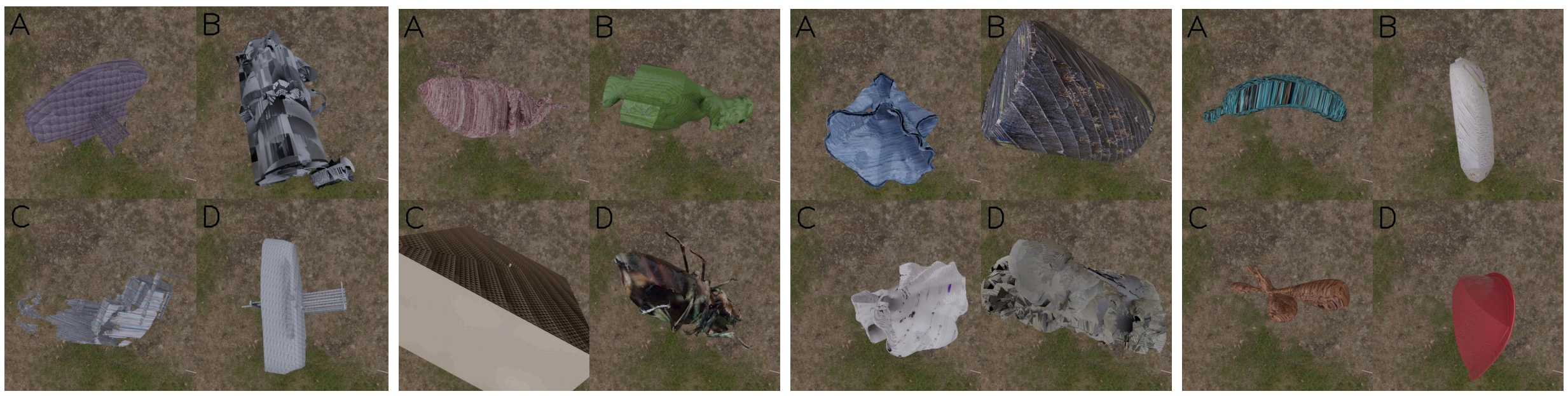

**Figure 5) Test 4: Which panel contains an object with a 3D shape identical to that in Panel A but with a different orientation and texture/material?**

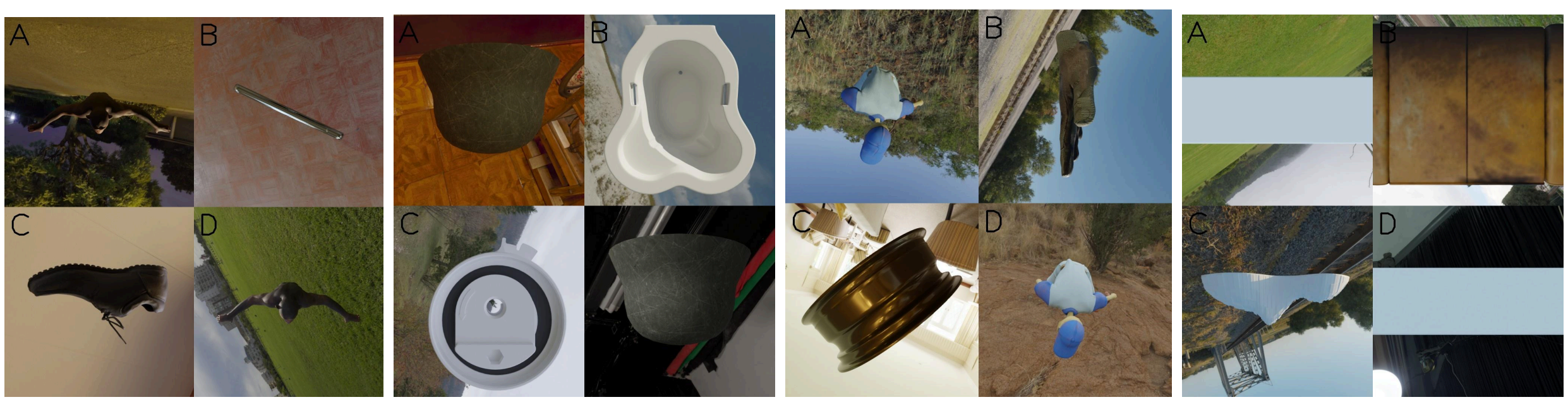

**Figure 6) Test 5: Which panel contains an object with a 3D shape identical to that in panel A but with a different background environment?**

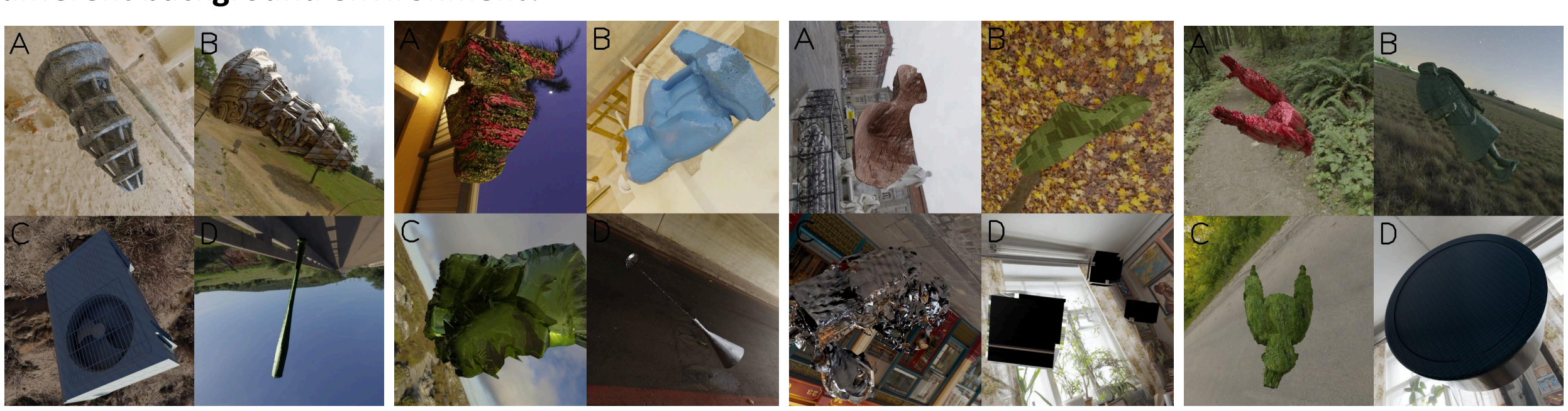

**Figure 7) Test 6: Which panel contains an object with a 3D shape identical to that in panel A but with a different orientation, environment, and texture/material?**

# 3. Testing the models

To test the model's ability to recognize 3D shapes we create a four-panel image (Figure 1-6). Two panels contain the same 3D shape but with some variation (orientation, texture, environment, Figures 2-6) and the model asks to find which panel contains an object with 3D shape identical to the object in panel A. The test images (Figure 1-6) were automatically made by randomly selecting four generated images (section 2). Two of the selected images are of the same 3D object but with one or more modifications (orientation/texture/environment) while the 2 other images are of different objects. The 4 images are arranged in a 2x2 grid (Figures 1-6), and a letter is added for each panel at the top left (A,B,C,D). Next, we give the model the image and ask it to identify which of the panels B,C,D contain an object with 3D shape identical to the shape of the object in panel A. Since both the testing and image generation are automatic this allows the creation of a large number of different images and tests. 1000 unique images were used for each test. Running the tests was done automatically using the models python API. The results are given in Table 1.

## 3.1. Prompt selection

An important aspect of the above test is how to express the question given to the model. While in theory, any clear description of the task should work, in practice different prompts can lead to up to 8% difference in accuracy. Interestingly detailed and exact prompts perform worse than short general ones. For example the short prompt:

***"Which of the panels contains an object with an identical 3D shape to the object in panel A. Your answer must come as a single letter"***

Give better accuracy than the long detailed prompt:

***"Carefully analyze the image. In panel A, there is an object with a specific shape. Your task is to identify which other panel (B, C, or D) contains an object that***
***1) Has the exact same 3d shape as the object in panel A.***
***2) Has a different orientation compared to the object in panel A.***
***3)Has a different texture compared to the object in panel A.***
***Respond with ONLY the letter of the panel (B, C, or D) that meets all these criteria."***

Specifically, pointing out the various possible changes seems to have negative effects. Asking the models to write their own prompts led to longer, more detailed text, which led to worse results. One explanation for this is that the models understand the task quite well from the short prompt and the longer prompt just consumes more attention. There are few strong indications that all models clearly understood the task. First when asked to explain their choices they clearly give explanations that are consistent with the tasks and the object in each panel (See Appendix).

Second the accuracy of the answers while below human is well above random (33%) and are higher and easier the task is (Table 1). We note models like Claude and LLama sometimes give long answers even when asked to only give panel letters, and sometimes refuse to answer claiming that no panel contains an identical 3D shape. In these cases, the question was repeated several times as is and then with increased assertiveness. While in case of a long answer the panel the model was asked to shorten the answer to one letter, we verified manually that this rarely adds error to the original answer. The results for the best prompts are shown in Table 1.

**Table 1: Result for 3D shape recognition: the top rows mark which property is changed between two images of the same 3D object. V means the property is changed. x means the property remains constant. *Keep Original Texture* means each object maintains its original texture and colors, meaning texture based recognition is possible. To view samples of each test see the related figure (top row).**

| | | | | | | | |
|---|---|---|---|---|---|---|---|
| Varied Orientations | x | x | V | V | x | V | V |
| Keep Original Texture | V | V | V | x | x | x | x |
| Varied Textures | x | x | x | x | V | V | V |
| Varied Backgrounds | x | V | x | x | x | x | V |
| Claude-3.5-sonnet | 98% | 97% | 93% | 82% | 88% | 74% | 69% |
| Gemini-1.5-pro | 100% | 98% | 99% | 87% | 93% | 83% | 81% |
| Gemini-2.0-Flash | 100% | 98% | 99% | 87% | 90% | 77% | 76% |
| GPT 4o | 100% | 98% | 97% | 89% | 96% | 85% | 82% |
| GPT 4 mini | 100% | 95% | 95% | 82% | 91% | 77% | 73% |
| LLama3.2-90b | 82% | 72% | 65% | 51% | 56% | 45% | 44% |
| Human | 100% | 100% | 100% | 98% | 98% | 98% | 97% |

# 4. Results

The results in Table 1 clearly show that all the models grasped the tasks and have some understanding of 3D shapes which is way above random (33%), even for cases where the object material is replaced or when viewed from a different angle. Gemini and GPT 4o clearly excel in this task but all models show some level of understanding (Table 1). At the same time, the performance of all models is still significantly lower than that of the average human. Suggesting the models achieve only a partial understanding of 3D shapes. Looking at Table 1 it seems that if either the texture or orientation remains the same the model can recognize the object with high accuracy. However if both the orientation and material are replaced the accuracy of all models drops dramatically. Replacing the background and illuminations (HDRI) has the smallest effect on accuracy, suggesting that none of the models have any challenge separating the object from its surroundings or accounting for different illumination effects.

When asked to explain their answers (Appendix) the models seem to do one of two responses: Offer the set of rotations and transformations leading from one image to another (Appendix). This suggests that the model has some kind of 3D representation of the object (or imagine it does[17]). However, even when the answers are right this transformation often seems made up and often refers to resizing the object and mirror image, although neither resizing nor mirroring was used. Another type of explanation given by the model refers to some set of 3D features like flat, dishlike, folded, etc, or to the object class. At this level, it seems that the models clearly understand the object in each image although these features are often too general to distinguish between different 3D shapes, especially for different objects of the same class. However, it's not clear how much the models actually can explain their own reasoning, even humans, for whom this task is relatively trivial will often find it hard to explain their decision in words. For humans, the results on all tests were nearly 100% although the fact that the images were generated automatically using a large number of assets leads to a small number of cases, where the images are unclear, too dark, or error technical issues in rendering process, this effect 1-2% of the images and explain the less than perfect human accuracy (Table 1).

# 5. Conclusion

The results of this work show that large vision models have gained an abstract understanding of 3D shapes, but still trail far beyond humans in this basic task. These results are consistent with previous works which show that despite their impressive performance Vision Language Models (VLM) often miss basic aspects of reality[8-12]. It might be expected that abstract 3D shape recognition will emerge from learning other tasks[16]. However, this apparently is not enough for current models. It is likely that this lack of basic 3D understanding can harm the models performance in many downstream tasks, and hence training these models directly for this task might be necessary. A major advantage of the synthetic data approach is that in addition to testing, it also allows for generating an unlimited amount of diverse data that can be used to train models on this task.

# 7. Supporting materials

The 3D shape Matching Dataset images are available at these URLs: GITHUB, ZENODO
Code used to generate the dataset is available at this URL.
Code used to evaluate the models on the benchmark available at this URL.

# Appendix: Sampled explanation given by models to justify their answers.

## Each response refers to the image above it.

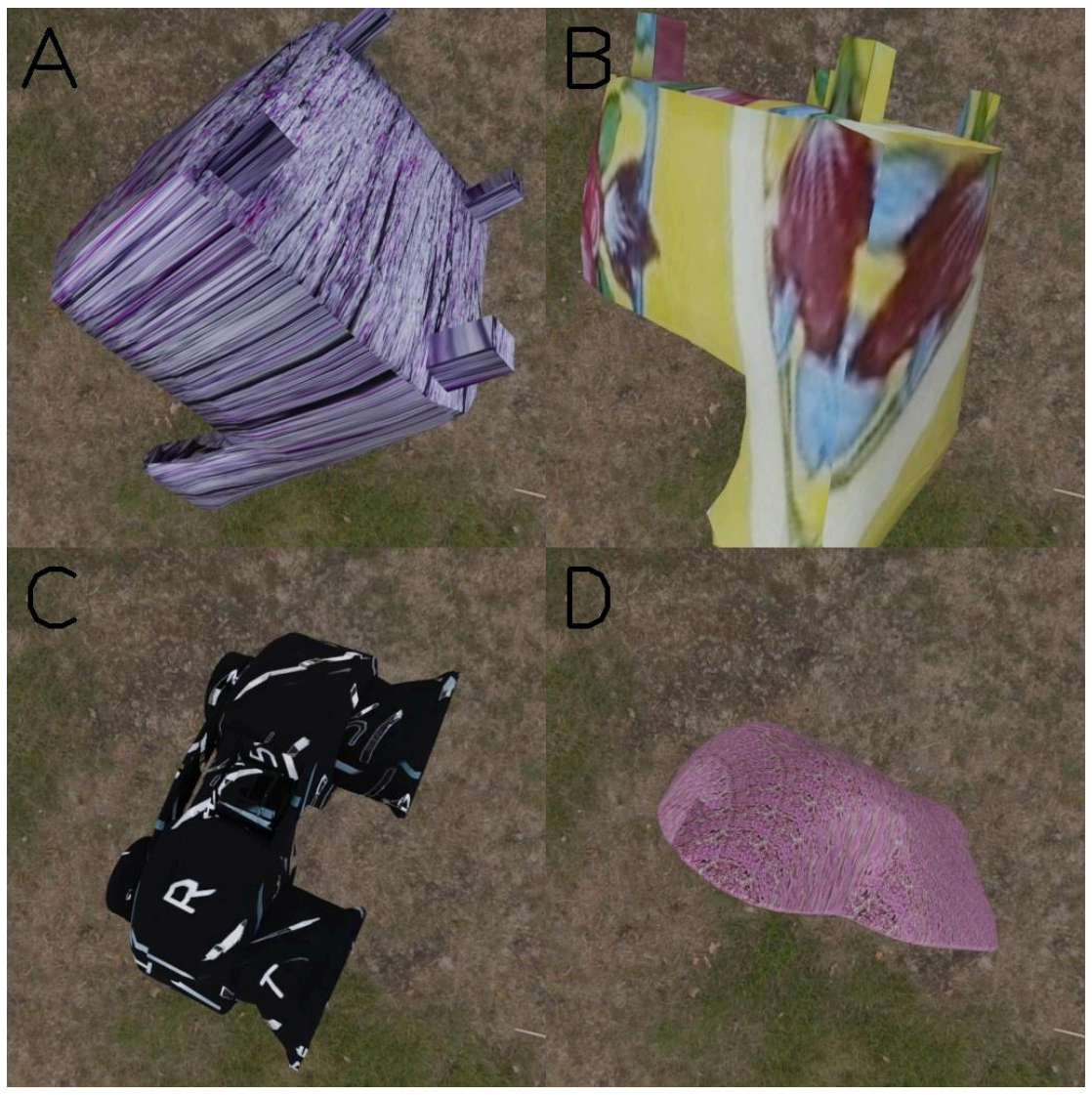

**Query: “Which panel contains an object that has identical 3D shape to the object in panel A, but different in orientation and texture. Explain”.** GPT 4o response: “Image D is the same shape as A, just rotated, smaller, and with a different texture. It's like taking the object from A, covering it with a pink, bubbly material, and then turning it over.”

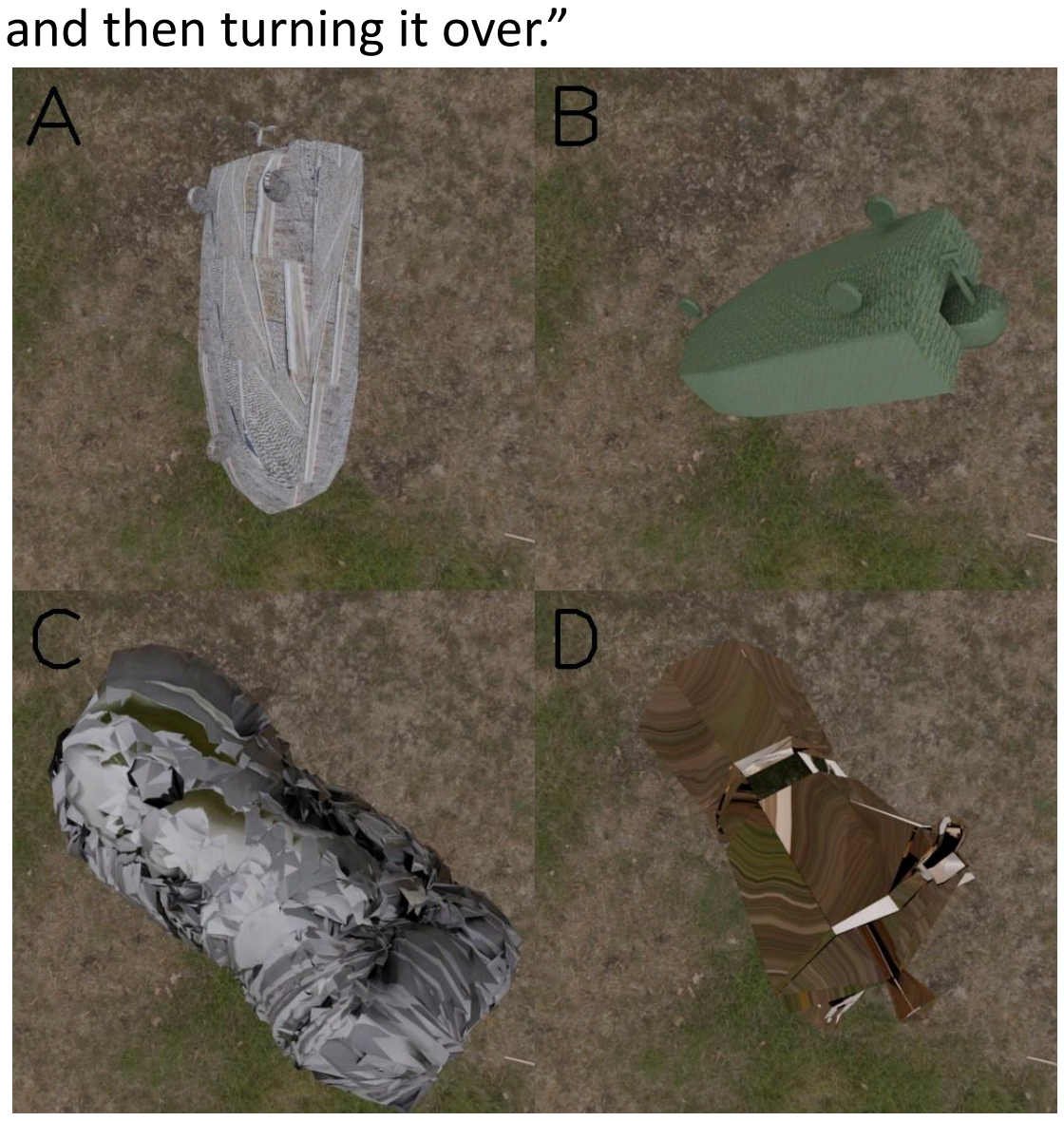

GPT 4o response: “GPT 4o response: Panel **D** is the same shape as A, just rotated and with a different texture. It appears to be made of polished wood, while A seems to be made of some sort of woven or layered material.”

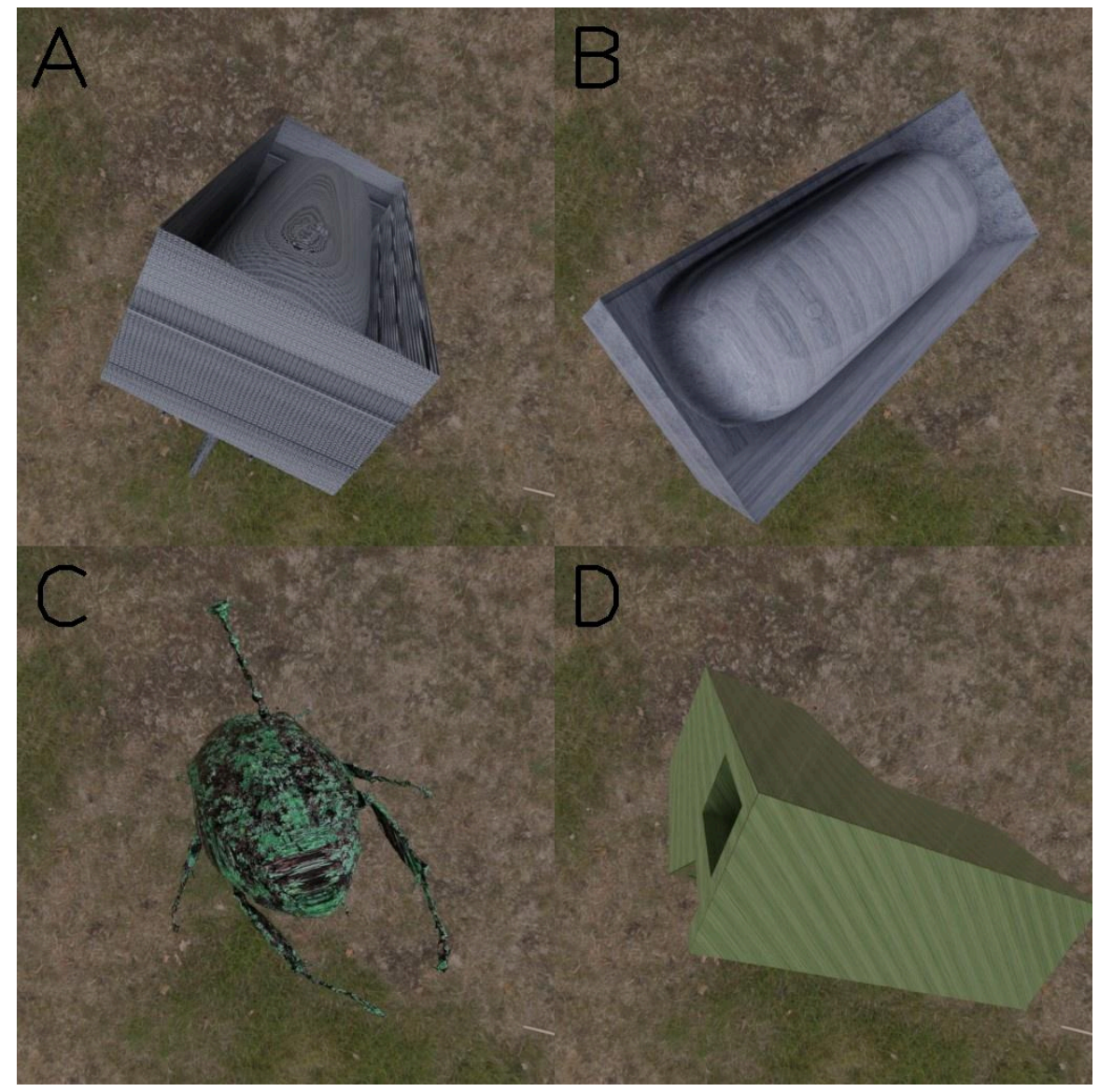

GPT 4o response: “Panel D is the match for A. It's the same basic triangular prism shape, just rotated and with a different (wood-like) texture applied.  Panel B is a rectangular prism with a curved indentation, and C is an insect”

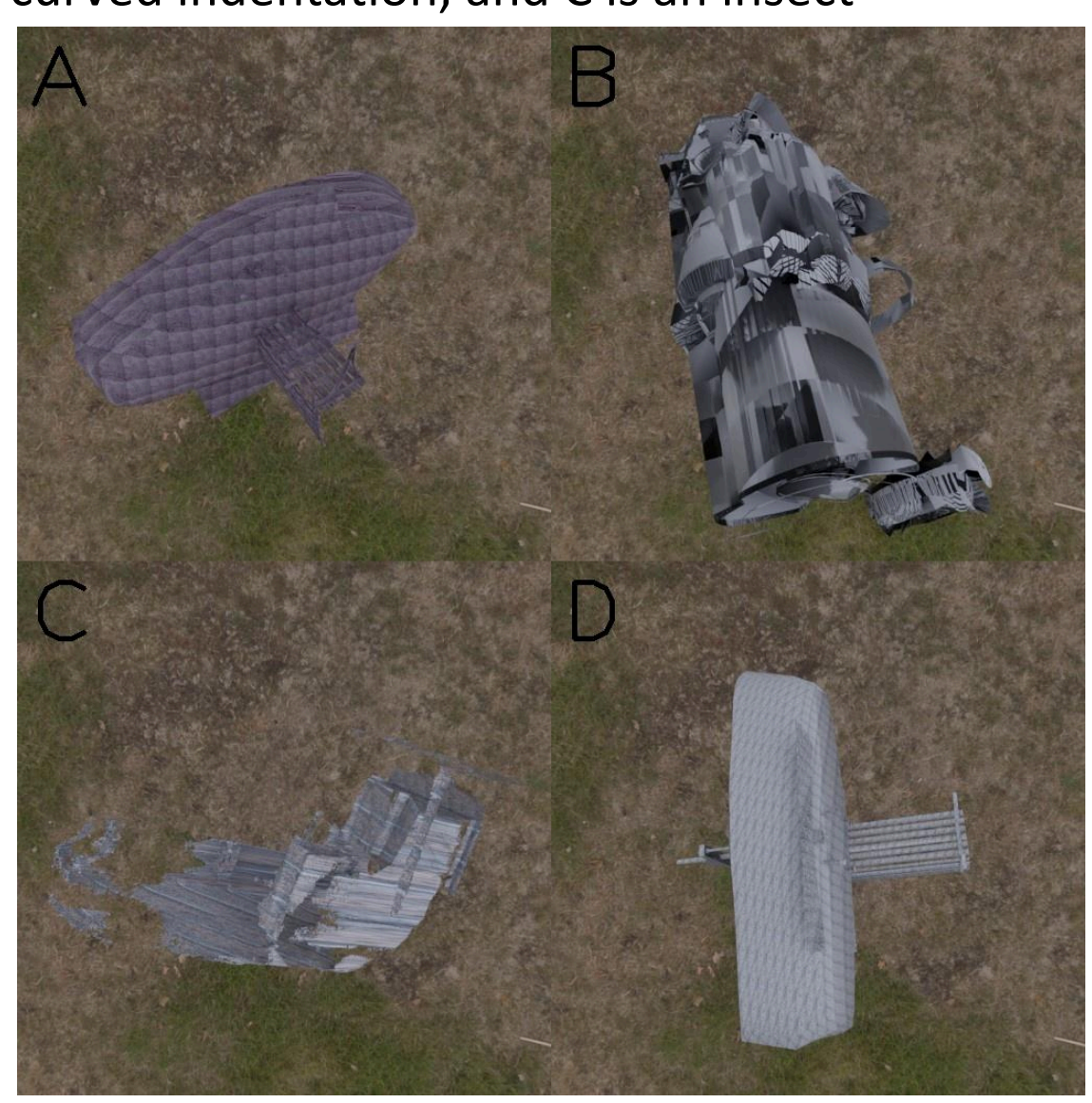

GPT 4o response (correct): “Panel **D** shows the object that matches panel A's shape.  It's clearly the same basic form (a rounded, dish-like object with an attached structure), but it's rotated and has a completely different texture/material appearance. It also seems to be mirrored from left to right.”

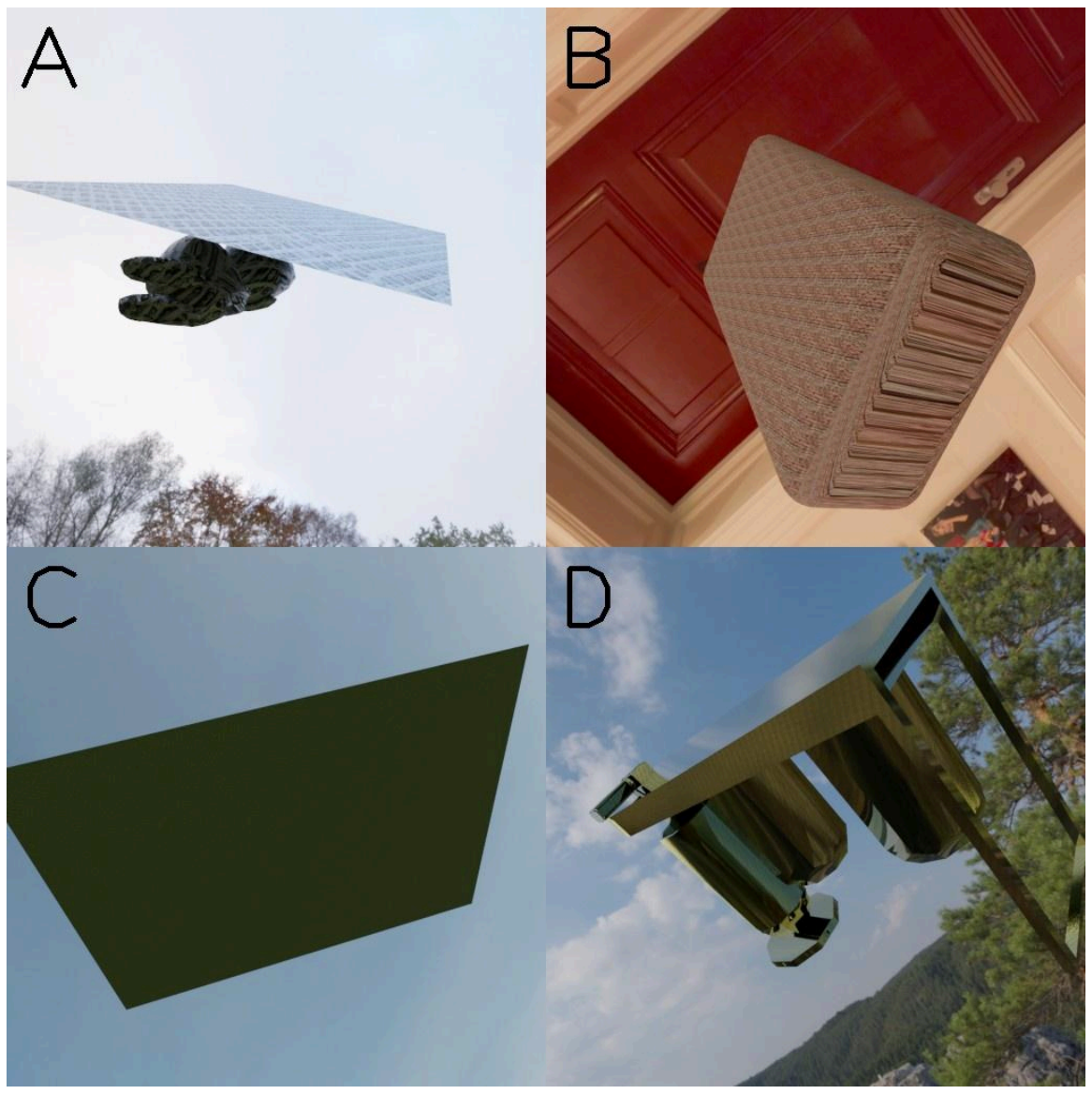

**Querry: "Which panel in the image contains an object similar in 3D shape to the object in panel A, explain".** Gemini response (Correct): "The object in panel C is similar in 3D shape to the object in panel A. Both objects resemble a flat or planar structure. While panel A shows the object from a slightly angled underside view, panel C shows a flat, dark shape without detailing, indicating a similar planar form."

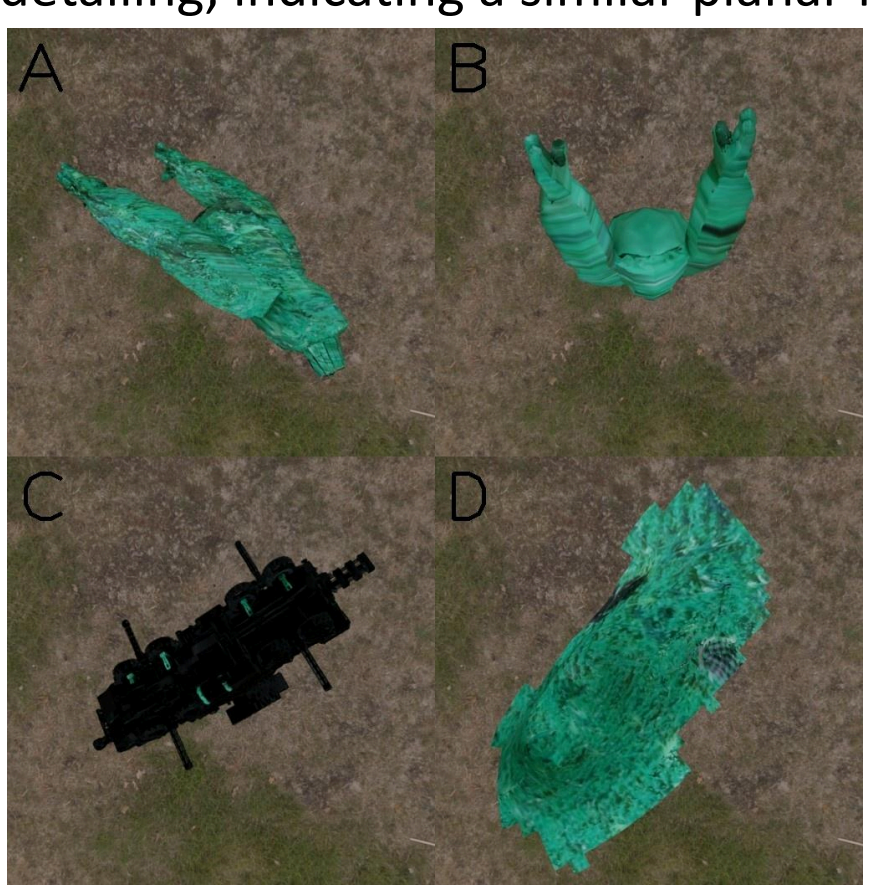

Gemini response " Certainly! Let's analyze the shapes in the image.
**Analysis:**
* **Panel A:** Shows a green, somewhat elongated, and distorted humanoid-like figure. It appears to be lying on the ground.
* **Panel B:** Shows a green humanoid-like figure standing straight with its arms raised. It is similar to A but not identical, as it's upright and the pose is different.
* **Panel C:** Shows a black metallic object, likely a machine or weapon, which bears no resemblance to the other objects.
* **Panel D:** Shows a green shape that appears to be a distorted version of the same humanoid-like figure as in panel A. It is leaning slightly.

**Conclusion:**

The panel that contains an object with an identical 3D shape to the object in panel A is **panel D**.
**Reasoning:**
Panel D's object clearly has the same structure and texture as panel A, it is only seen from a different angle and it appears to have rotated and tilted compared to panel A. Panel B and C are distinctly different objects and shapes."

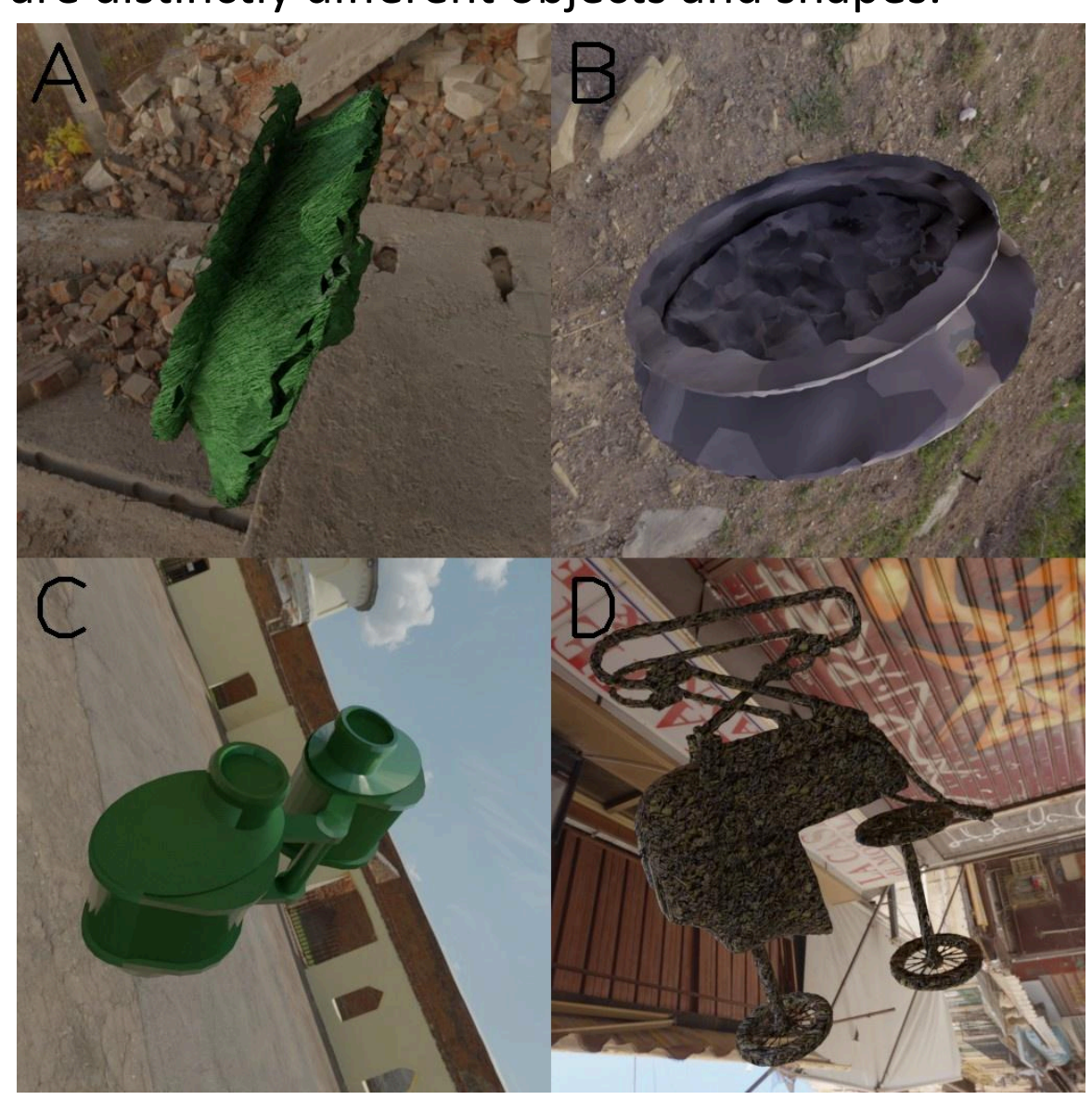

Gemini's response: "The object in panel A appears to be an irregular, elongated shape. The panel containing a similar 3D shape is likely panel D. Both objects have elongated, complex structures, though they are not identical.

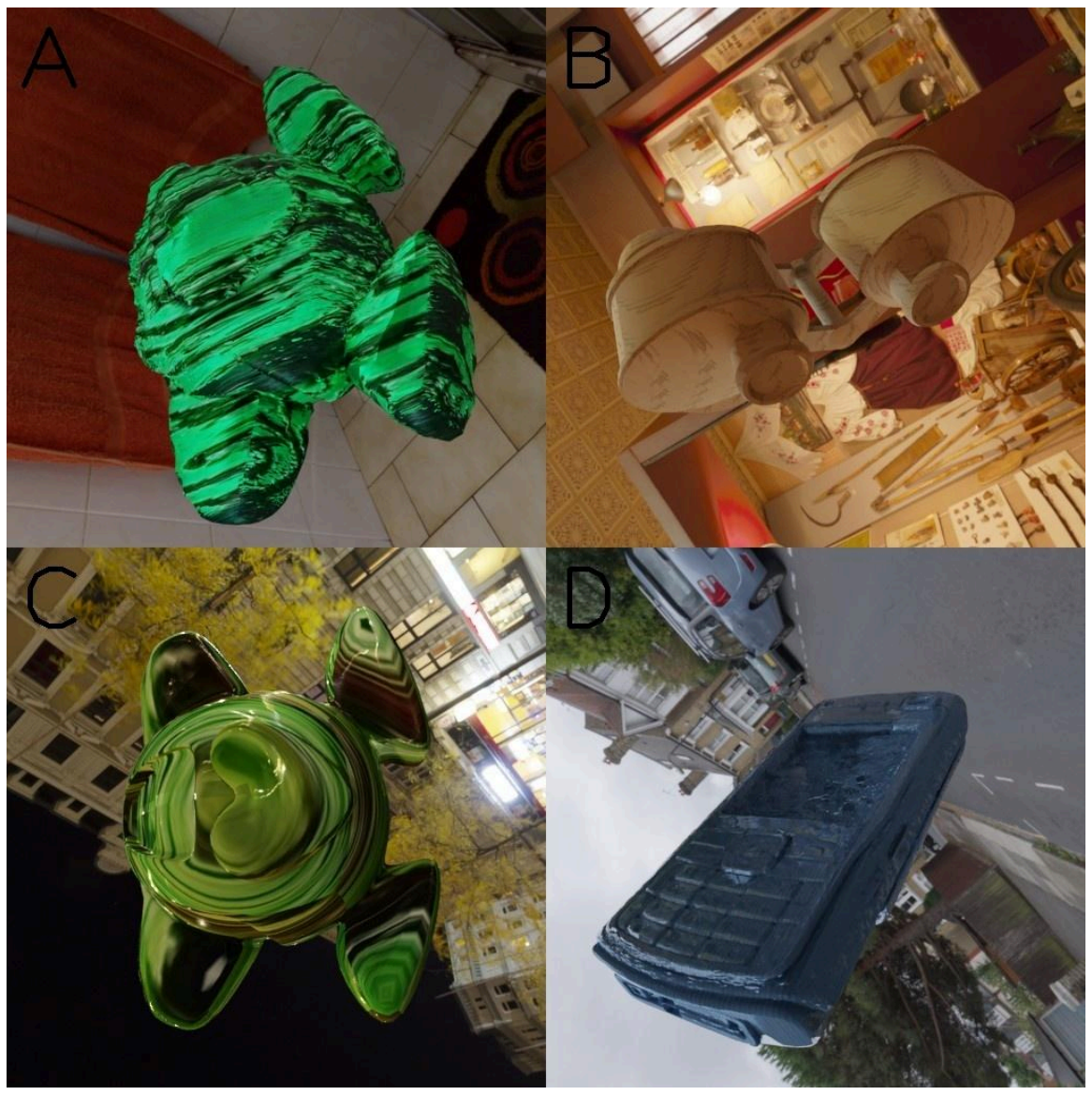

**Query: "Which of the panels contain an object with identical 3d shape to the object in panel A? Explain your answer."**

Claude Response: " Looking at all panels:

Panel A shows a shiny green balloon-like or inflated object with a lumpy, irregular shape.

Panel B shows mushroom-like shapes in beige/white colors.

Panel C shows a shiny green figure that appears to be a cartoon character or mascot sculpture with rounded shapes and ears.

Panel D shows what appears to be a dark-colored vehicle or bus from an unusual angle.

None of the other panels (B, C, or D) contain objects with identical 3D shapes to the irregular, lumpy form shown in Panel A. While Panel C is also green and shiny, it has a very different, more structured and character-like form compared to the abstract, bulbous shape in Panel A. Panels B and D show completely different types of objects with very different geometries.

Therefore, none of the other panels contain objects with identical 3D shapes to Panel A."

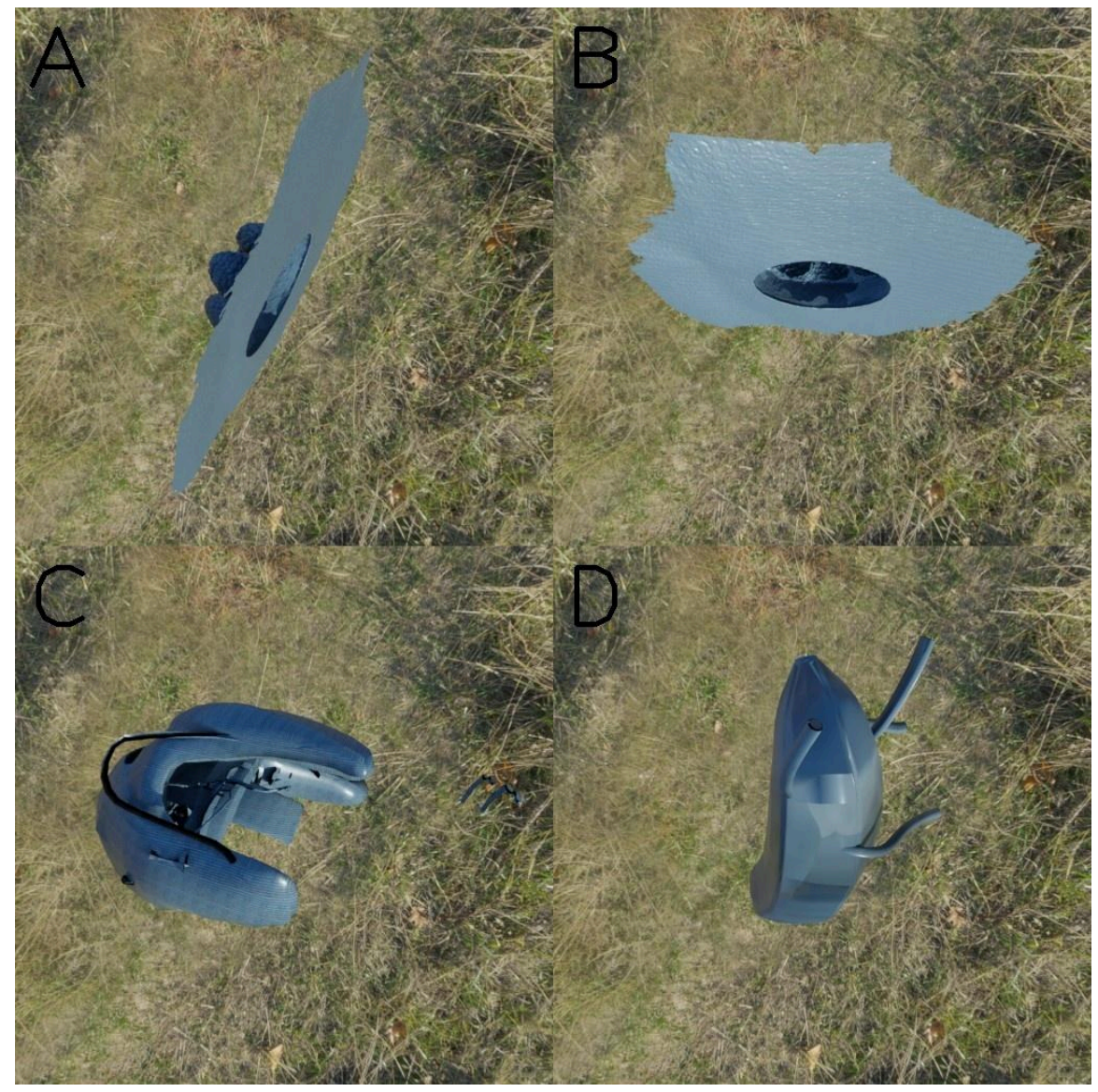

**Query: "Which of the panels contain an object with identical 3d shape to the object in panel A? Explain your answer."**

Claude Response: " Let me analyze each panel:

Panel A shows a narrow, elongated object with a gray/blue coloring.

Looking at the other panels:

- Panel B shows a wider, more rounded shape that appears quite different from A
- Panel C shows a curved, U-shaped object that has a very different form than A
- Panel D shows an object that appears to be the same as A but from a different angle, maintaining the same elongated form and proportions

Therefore, Panel D contains the object with identical 3D shape to Panel A. They appear to be the same object photographed from different viewpoints - Panel A shows a side view while Panel D shows a perspective view, but the fundamental 3D geometry remains the same. Panels B and C show distinctly different shapes that could not be the same object viewed from different angles.

The answer is Panel D."

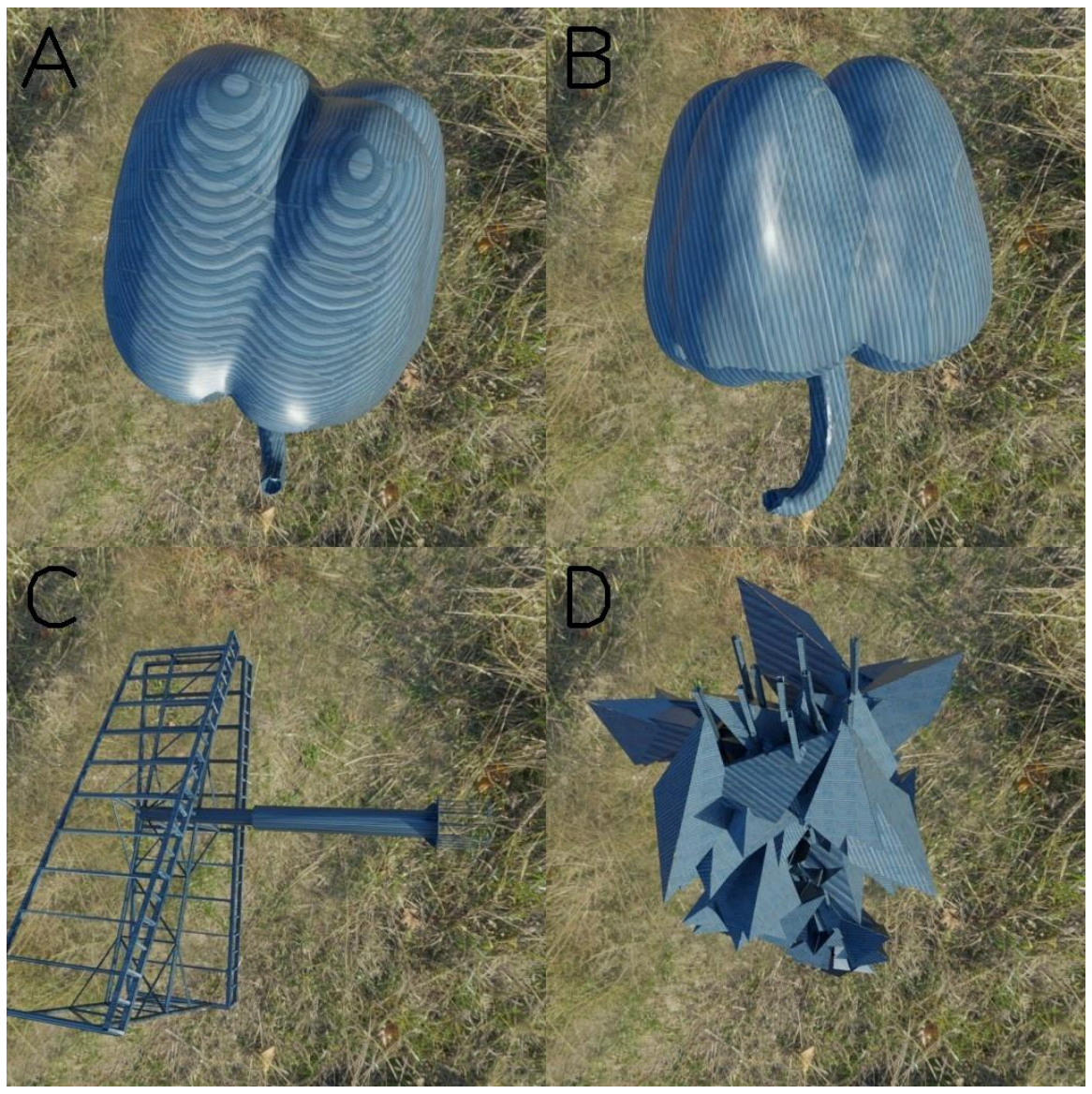

**Query "Which of the panels contain an object with identical 3d shape to the object in panel A? Explain your answer."** LLAma response: "
**Identical 3D Shapes in Panels B and C**

Upon examining the image, it is evident that panels B and C contain objects with identical 3D shapes as the object in panel A.

**Reasoning:**
* The object in panel A has a distinctive shape with a curved top and a narrow base.
* Panel B displays a similar object with the same curved top and narrow base, indicating that they share the same 3D shape.
* In panel C, the object is oriented differently, but its curved top and narrow base are still visible, confirming that it also shares the same 3D shape as the object in panel A.

**Conclusion:**

Based on the visual evidence, panels B and C contain objects with identical 3D shapes as the object in panel A."

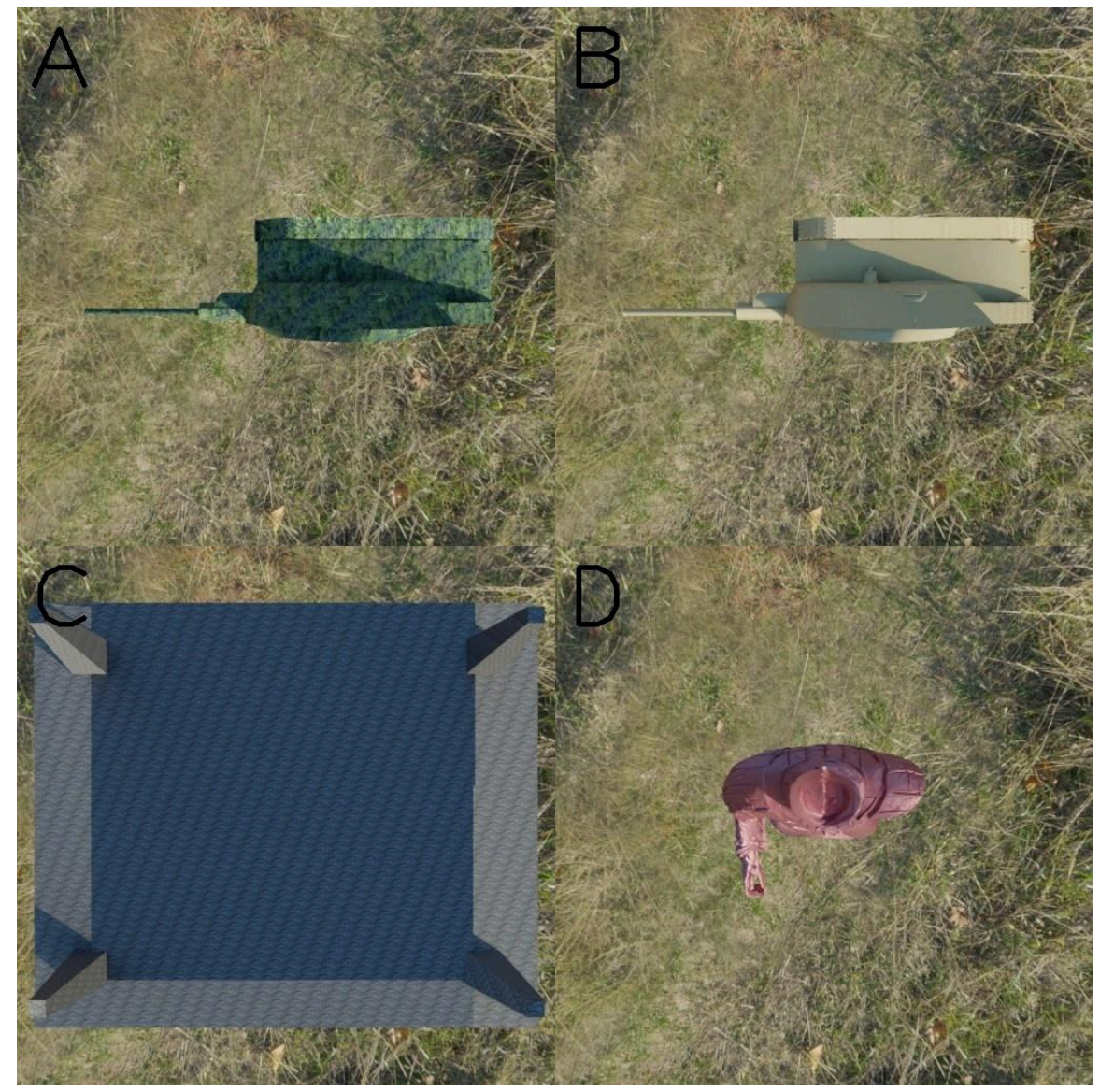

**Query "Which of the panels contain an object with identical 3d shape to the object in panel A? Explain your answer."** LLAma response: "
The object in panel A is a tank. The object in panel D is also a tank, but it appears to be a different model or type of tank. The object in panel B is not a tank, but rather a piece of artillery or a cannon. The object in panel C is not a tank, but rather a building or a structure.
Therefore, the panels that contain an object with an identical 3D shape to the object in panel A are panels A and D.

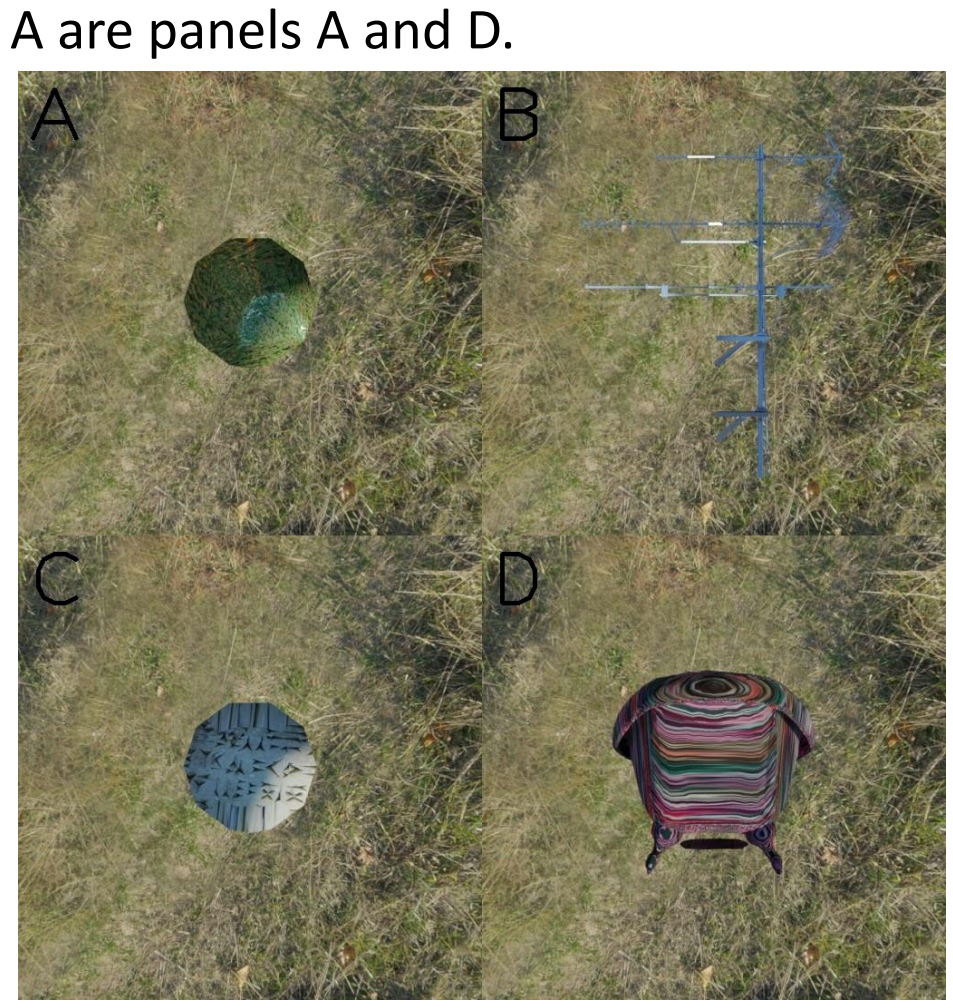

LLama response: Panel A shows a three-dimensional object that resembles an octahedron. The same object appears in Panels C and D. However, Panel B shows an abstract blue figure with arrows and white bars, so it is not real. Answer: Panels A and D

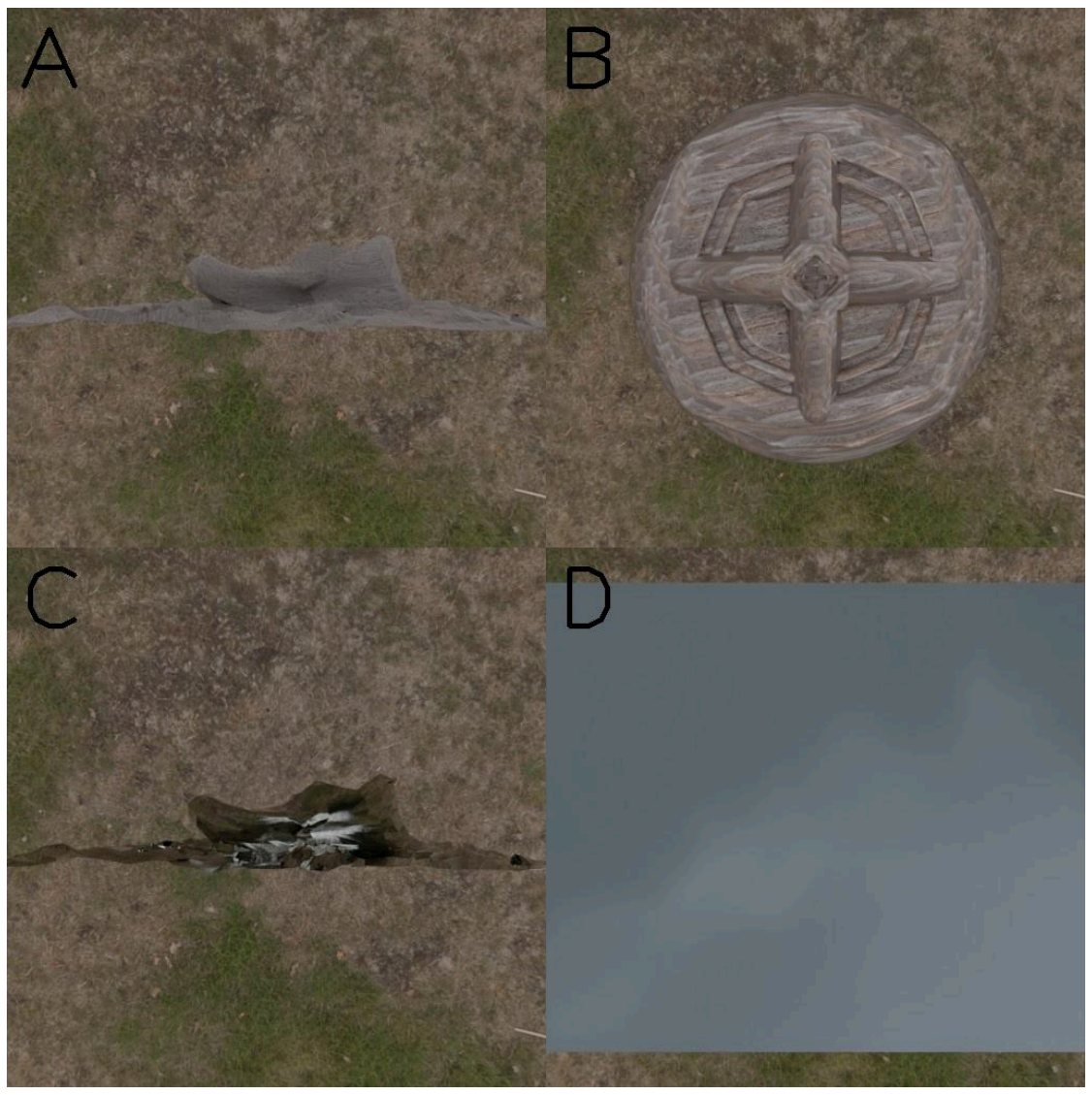

**Query: "Which of the panels contains an object with an identical 3d shape to the object in panel A but with a different texture/material?"** Claude's response: "B - The shape in panel B appears to be the same draped/folded form as in panel A, but with a different texture that has a carved/embossed cross or compass-like pattern on it, while panel A shows a plain gray material."

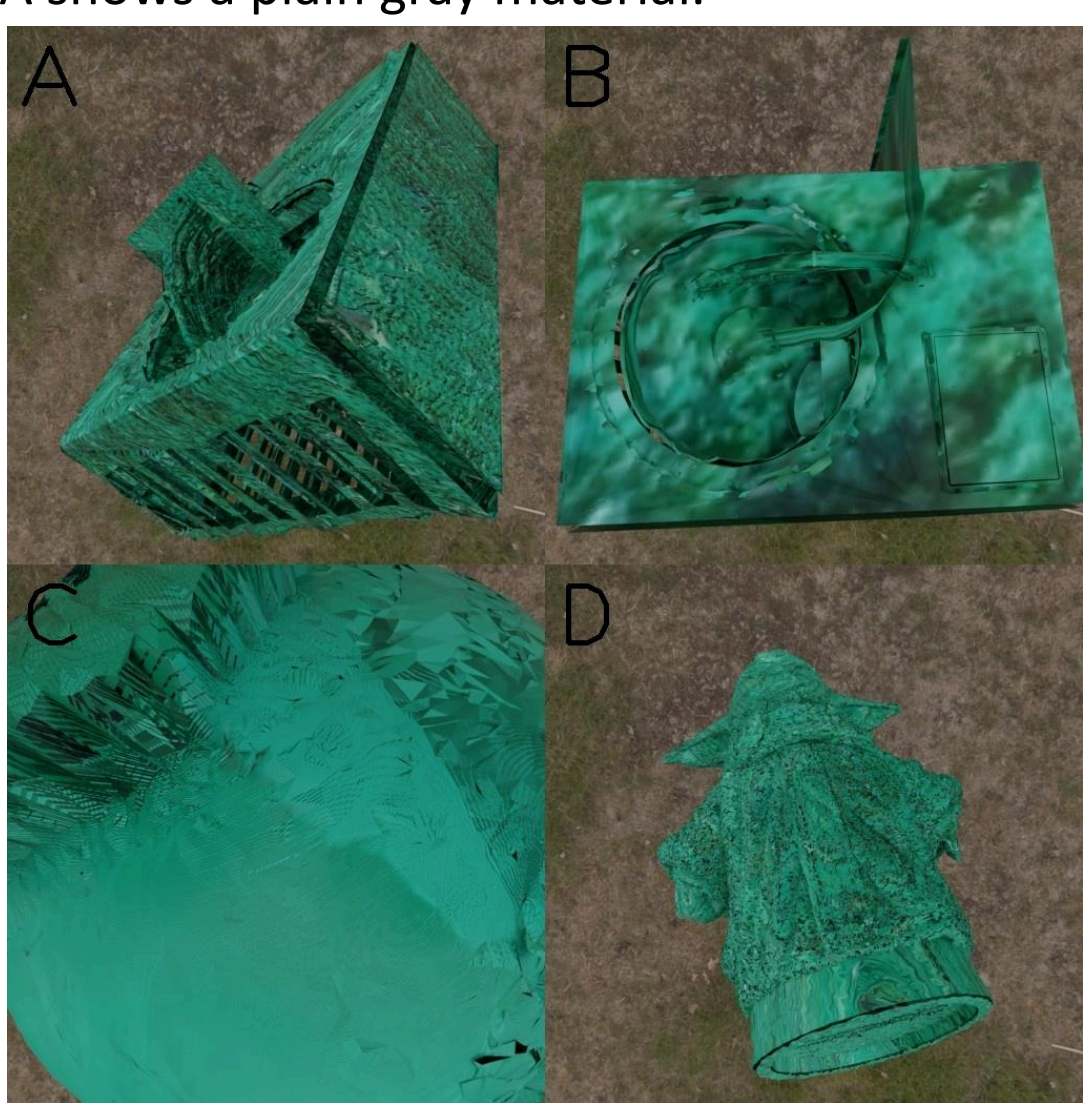

Gemini Response: " Based on the images provided, none of the panels (B, C, or D) contain an object with the exact same 3D shape as the object in panel A.
**Explanation:**
* **Panel A:** Depicts what appears to be a rectangular building structure with a slanted roof section. There are also visible windows in the structure. The surface looks rough and textured.
* **Panel B:** Shows an object with a complex circular structure and a vertical element, resembling a sundial or similar device. It's quite different from the rectangular shape in A.
* **Panel C:** Presents a round, dome-like object. The surface is textured, but there are no straight edges, unlike panel A.
* **Panel D:** Depicts what appears to be a standing figurine with a cloak and pointed ears, likely based on the character Yoda. This is completely different in form to the building of panel A. **In summary:** The key to determining that there aren't identical 3D shapes is that while they share similar textures or colors (green), the basic forms of the objects in panels B, C, and D are significantly different from the rectangular building with a slanted roof as seen in panel A."

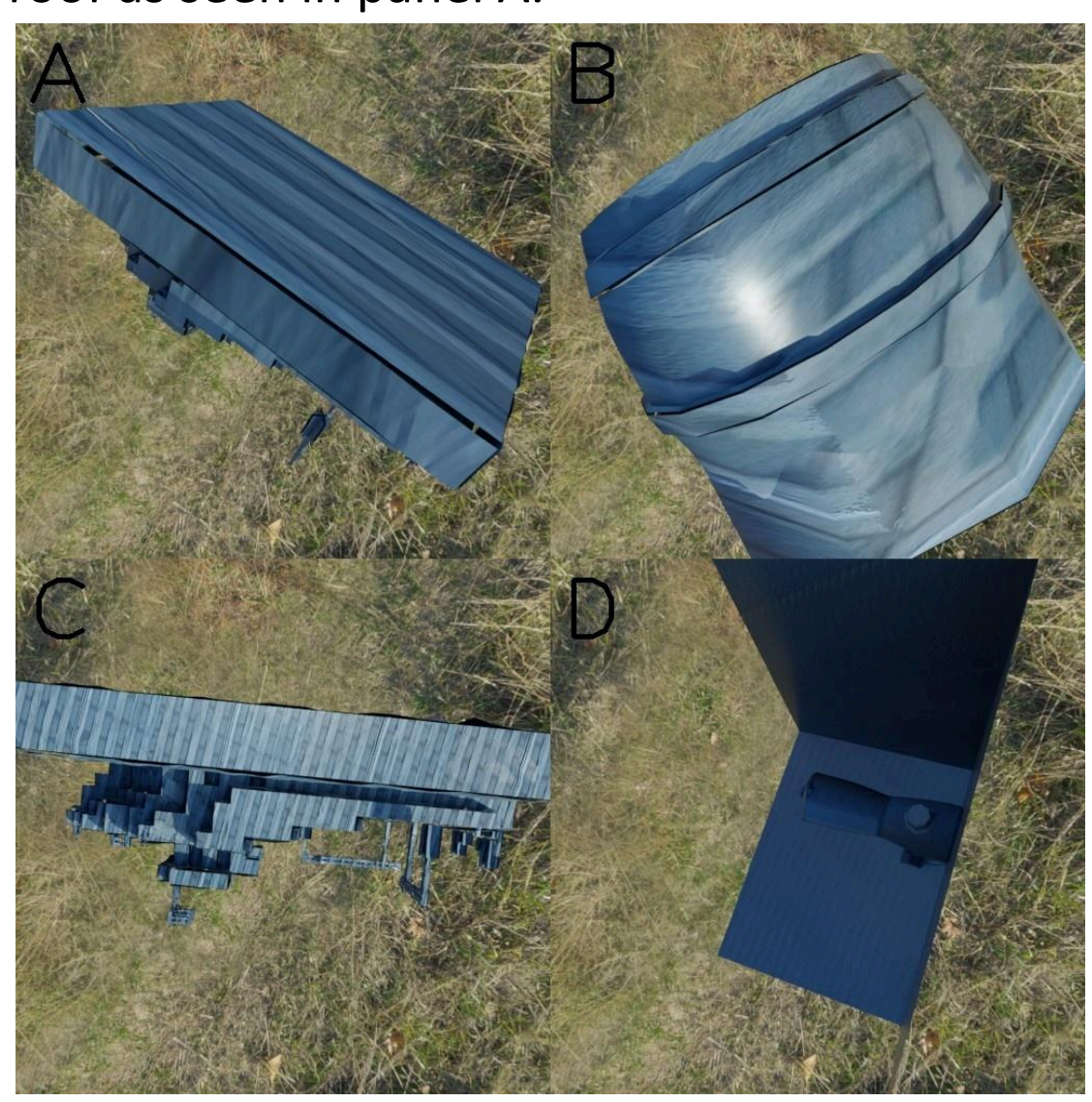

GPT 4O Response: "The object in panel D has an identical 3D shape to the object in panel A. Both objects have similar flat, elongated rectangular forms without any visible extrusions or uneven surfaces. The other panels contain objects with distinct shapes, such as the cylindrical form in panel B and the jagged, irregular form in panel C."